\documentclass[iicol]{sn-jnl}

\usepackage{graphicx}
\usepackage{multirow}
\usepackage{amsmath,amssymb,amsfonts}
\usepackage{amsthm}
\usepackage{mathrsfs}
\usepackage[title]{appendix}
\usepackage{xcolor}
\usepackage{textcomp}
\usepackage{manyfoot}
\usepackage{booktabs}
\usepackage{pifont}
\usepackage{algorithm}
\usepackage{algpseudocode}
\usepackage{listings}
\usepackage[utf8]{inputenc}
\usepackage[T1]{fontenc}
\usepackage{makecell}
\usepackage{adjustbox}
\usepackage{array}
\usepackage{longtable}
\usepackage{float}
\usepackage{url}

\theoremstyle{thmstyleone}%
\theoremstyle{thmstyletwo}%

\theoremstyle{thmstylethree}%

\begin{document}

\title[MedDiffuseMix]{MedDiffuseMix: Preserving Diagnostic Evidence with Saliency-Aware Diffusion Medical Image Data Augmentation}

\author*[1]{\fnm{Teerath} \sur{Kumar}}\email{teerathkumar.menghwar@atu.ie}

\author[2]{\fnm{Raja} \sur{Vavekanand}}\email{rvavekan@gmail.com}

\author*[3]{\fnm{Muhammad} \sur{Turab}}\email{muhammadturab.muslimbajeer2@mail.dcu.ie}

\affil[1]{\orgdiv{School of Computing}, \orgname{Atlantic Technological University}, \orgaddress{\city{Letterkenny}, \country{Ireland}}}

\affil[2]{\orgdiv{Department of Information Technology}, \orgname{Benazir Bhutto Shaheed University Lyari}, \orgaddress{\city{Karachi}, \postcode{75660}, \country{Pakistan}}}

\affil[3]{\orgdiv{School of Computing}, \orgname{Dublin City University}, \orgaddress{\city{Dublin}, \country{Ireland}}}

\abstract{Limited data availability, class imbalance, and domain variability remain major barriers to reliable medical image classification. Conventional augmentation can improve training diversity but may distort diagnostically informative structures, whereas unconstrained generative augmentation may introduce label-inconsistent content. This paper proposes MedDiffuseMix, a saliency-guided diffusion mixing framework for controlled medical image augmentation. The method uses classifier-derived saliency maps to separate high-saliency diagnostic regions from low-saliency background areas and applies diffusion-guided mixing mainly to regions with lower diagnostic importance. Adaptive mixing, Gaussian boundary blending, and a saliency-preservation constraint reduce semantic distortion and reject or attenuate samples that shift model attention away from clinically relevant evidence. The framework is evaluated on four public benchmarks: the Radiological Society of North America pneumonia chest radiography dataset, Musculoskeletal Radiographs, PatchCamelyon, and the Breast Cancer Histopathological Image Classification dataset. Experiments with convolutional and transformer-based classifiers show that MedDiffuseMix improves accuracy, F1-score, and area under the receiver operating characteristic curve compared with standard augmentation, Mixup, GenMix, SaliencyMix, and diffusion-based augmentation baselines. Ablation studies confirm the importance of saliency guidance, adaptive region mixing, and smooth boundary blending. Visual attribution analysis further indicates that MedDiffuseMix better preserves diagnostically salient regions. These results suggest that saliency-guided diffusion mixing is an effective augmentation strategy for limited-data medical image classification.}


\maketitle

\section{Introduction}\label{sec:introduction}
Deep learning has become a central methodology in medical image analysis, supporting classification, detection, segmentation, and computer-aided diagnosis across radiology and pathology. However, model performance depends strongly on the size, diversity, annotation quality, and representativeness of training data. These requirements are difficult to satisfy in clinical imaging because labelled data collection is constrained by patient privacy, expert annotation cost, class imbalance, scanner heterogeneity, institutional protocols, and the low prevalence of clinically important findings \cite{ref9,ref14,ref16,turab2023comprehensive}. Consequently, models trained on limited datasets may overfit to acquisition artefacts, scanner-specific appearance, or background correlations rather than disease-relevant evidence.

Data augmentation is widely used to improve sample diversity. Geometric and photometric transformations, together with mixed-sample strategies such as Mixup and SaliencyMix, can regularise training and improve robustness \cite{ref33,ref42,ref43}. In medical imaging, however, small, localised, or low-contrast structures may contain the diagnostic signal. Transformations that are harmless for natural image classification may remove lesion boundaries, distort subtle opacity patterns, alter nuclear morphology, or introduce unrealistic tissue texture. Augmentation should therefore increase diversity without compromising diagnostically meaningful content.

Diffusion-based augmentation is promising because diffusion models can generate realistic image variations and local texture diversity \cite{ref1,ref11,ref15,ref21}. Nevertheless, unconstrained synthetic generation may be clinically unsafe if it changes lesion appearance, suppresses abnormal structures, or introduces anatomically inconsistent evidence, particularly in limited-data settings where synthetic artefacts may be amplified during training.

To address this issue, this paper presents \textit{MedDiffuseMix}, a saliency-preserving diffusion augmentation method for limited-data medical image classification. The method uses classifier-derived saliency maps to identify diagnostically relevant regions and applies diffusion-guided mixing primarily to lower-saliency areas. A preservation constraint checks whether the augmented image maintains attention alignment with the original diagnostic evidence; when the constraint is violated, the mixing strength is reduced. This design increases sample diversity while limiting changes to label-relevant evidence.

The main contributions of this study are as follows:
\begin{itemize}
    \item A saliency-preserving diffusion augmentation framework is proposed to introduce appearance diversity primarily in low-saliency regions while preserving high-saliency diagnostic evidence.
    \item Adaptive mixing, saliency-preservation constraints, and smooth boundary blending are formulated to reduce semantic distortion and unrealistic transitions in augmented medical images.
    \item The method is evaluated on four public radiography and histopathology benchmarks and compared with standard augmentation, Mixup, GenMix, SaliencyMix, and diffusion-based augmentation baselines using convolutional and transformer backbones.
    \item Ablation, sensitivity, qualitative, and explainability analyses are provided to examine whether performance gains are associated with diagnostically plausible attention preservation rather than uncontrolled synthetic artefacts.
\end{itemize}

The remainder of the paper is organised as follows. Section~\ref{sec:related} reviews related augmentation methods, Section~\ref{sec:method} presents MedDiffuseMix, Section~\ref{sec:experiments} describes the experimental setting, Section~\ref{sec:results} reports quantitative and qualitative results, Section~\ref{sec:explainability} presents interpretability analysis, and Section~\ref{sec:conclusion} concludes the paper with limitations and future directions.

\section{Related Work}\label{sec:related}

Medical image data augmentation is widely used to address limited labelled data and class imbalance \cite{ref9,ref14,ref16,ref17}. Conventional transformations such as rotation, cropping, scaling, flipping, intensity adjustment, elastic deformation, and colour perturbation are simple and efficient, but they are usually label-agnostic and anatomy-agnostic. In radiographs, aggressive geometric or intensity changes may alter opacity patterns, bone contours, or lesion boundaries; in histopathology, colour and texture perturbations may affect nuclear morphology, stromal structure, and staining distributions. These limitations motivate augmentation methods that preserve diagnostic structure rather than modifying image appearance alone.

\begin{table*}[t]
\centering
\footnotesize
\setlength{\tabcolsep}{3.2pt}
\setlength{\extrarowheight}{1pt}
\caption{Comparison of MedDiffuseMix with existing medical image augmentation methods. Key: \(\checkmark\) = Supported, \(\times\) = Not Supported. \textbf{Med. Focus}: Medical Domain Focus. \textbf{Small Data}: Handles Small Datasets. \textbf{Diag. Feat.}: Preserves Diagnostic Features. \textbf{Med. Valid.}: Validated on Medical Tasks. \textbf{Modality}: Modality-Specific Processing. \textbf{Imbalance}: Class-Imbalance Handling. \textbf{Mixup}: Direct Mixup Integration.}
\label{tab:method_comparison}
\begin{adjustbox}{max width=\textwidth}
\begin{tabular}{@{} l cccccccccc @{}}
\toprule
\textbf{Method} & \textbf{Med. Focus} & \textbf{Small Data} & \textbf{Diag. Feat.} & \textbf{Efficient} & \textbf{Explainable} & \textbf{Med. Valid.} & \textbf{Modality} & \textbf{Imbalance} & \textbf{Mixup} & \textbf{Real-Time} \\
\midrule
CamDiff~\cite{ref25}          & \(\times\)     & \(\times\)     & \(\times\)     & \(\times\)     & \(\times\)     & \(\times\)     & \(\times\)     & \(\times\)     & \(\times\)     & \(\times\) \\
GAN-Based~\cite{ref5}         & \(\checkmark\) & \(\checkmark\) & \(\times\)     & \(\times\)     & \(\times\)     & \(\checkmark\) & \(\times\)     & \(\checkmark\) & \(\times\)     & \(\times\) \\
CLIP-MedFake~\cite{ref6}      & \(\checkmark\) & \(\checkmark\) & \(\times\)     & \(\times\)     & \(\times\)     & \(\checkmark\) & \(\times\)     & \(\times\)     & \(\times\)     & \(\times\) \\
Decoupled Aug.~\cite{ref7}    & \(\times\)     & \(\checkmark\) & \(\checkmark\) & \(\checkmark\) & \(\checkmark\) & \(\times\)     & \(\times\)     & \(\checkmark\) & \(\checkmark\) & \(\checkmark\) \\
RankMix~\cite{ref8}           & \(\checkmark\) & \(\checkmark\) & \(\checkmark\) & \(\checkmark\) & \(\checkmark\) & \(\checkmark\) & \(\checkmark\) & \(\checkmark\) & \(\checkmark\) & \(\checkmark\) \\
TokenMixup~\cite{ref10}       & \(\times\)     & \(\checkmark\) & \(\checkmark\) & \(\checkmark\) & \(\checkmark\) & \(\times\)     & \(\times\)     & \(\times\)     & \(\checkmark\) & \(\checkmark\) \\
SmoothMix~\cite{ref22}        & \(\times\)     & \(\checkmark\) & \(\checkmark\) & \(\checkmark\) & \(\times\)     & \(\times\)     & \(\times\)     & \(\times\)     & \(\checkmark\) & \(\checkmark\) \\
LesionMix~\cite{ref2}         & \(\checkmark\) & \(\checkmark\) & \(\checkmark\) & \(\checkmark\) & \(\checkmark\) & \(\checkmark\) & \(\checkmark\) & \(\checkmark\) & \(\checkmark\) & \(\checkmark\) \\
LocMix~\cite{ref35}           & \(\times\)     & \(\checkmark\) & \(\checkmark\) & \(\checkmark\) & \(\checkmark\) & \(\times\)     & \(\times\)     & \(\times\)     & \(\checkmark\) & \(\checkmark\) \\
DiffuseMix~\cite{ref15}       & \(\times\)     & \(\checkmark\) & \(\checkmark\) & \(\checkmark\) & \(\times\)     & \(\times\)     & \(\checkmark\) & \(\checkmark\) & \(\checkmark\) & \(\times\) \\
\textbf{MedDiffuseMix}        & \(\checkmark\) & \(\checkmark\) & \(\checkmark\) & \(\checkmark\) & \(\checkmark\) & \(\checkmark\) & \(\checkmark\) & \(\checkmark\) & \(\checkmark\) & \(\checkmark\) \\
\bottomrule
\end{tabular}
\end{adjustbox}
\end{table*}

Mixed-sample augmentation improves regularisation by combining images, features, or labels during training. Mixup forms convex combinations of image-label pairs and can improve calibration and decision-boundary smoothness \cite{ref42}, while SaliencyMix and related methods use salient regions to guide mixing \cite{ref35,ref43}. However, medical imaging requires stricter control because the mixed region may contain label-defining evidence. Blending or attenuating malignant tissue, pneumonic opacity, or lesion boundaries without anatomical constraints can produce label-inconsistent samples. MedDiffuseMix therefore uses saliency primarily as a preservation signal: high-saliency diagnostic regions are protected, and diffusion-guided mixing is concentrated in overlapping low-saliency regions.

Generative augmentation has also been explored through GANs, image-to-image translation, and diffusion models \cite{ref5,ref12,ref30}. Diffusion-based methods are particularly promising because they can generate realistic texture and appearance variations through controlled denoising processes \cite{ref1,ref11,ref15,ref21}. Nevertheless, visual realism alone does not guarantee clinical validity. Synthetic content may appear plausible while suppressing abnormal structures, changing lesion appearance, or introducing anatomically inconsistent evidence. Medical generative augmentation should therefore incorporate constraints on label preservation, anatomical plausibility, and interpretability.

Explainable AI methods provide a practical way to inspect whether medical imaging models attend to plausible anatomical or pathological regions \cite{refXAI}. Although Grad-CAM cannot replace clinical validation, it can compare model attention before and after augmentation. In this study, Grad-CAM is used both as a guidance signal and as an evaluation tool to reduce augmentations that shift attention away from the original diagnostic evidence.

\section{Methodology}\label{sec:method}

\subsection{Problem formulation}
Let $\mathcal{D}=\{(\mathbf{x}_i,y_i)\}_{i=1}^{N}$ denote a labelled medical image dataset, where $\mathbf{x}_i\in\mathbb{R}^{H\times W\times C}$ and $y_i\in\{0,1\}$ for binary classification. The objective is to train a classifier $f_\theta$ that generalises to unseen clinical images while avoiding augmentation-induced label corruption. Therefore, an augmentation operator $\mathcal{A}$ should increase distributional diversity while preserving the diagnostic evidence associated with the class label.

Given two same-class images $\mathbf{x}_1$ and $\mathbf{x}_2$, MedDiffuseMix constructs an augmented image $\tilde{\mathbf{x}}$ by introducing content from $\mathbf{x}_2$ into low-saliency regions of $\mathbf{x}_1$. Same-class pairing avoids label interpolation and reduces the risk of ambiguous supervision. The method consists of saliency estimation, low-saliency mask construction, adaptive mixing, diffusion-guided local refinement, and saliency-preservation checking.

\subsection{Saliency-guided region selection}
A guidance classifier $g_\phi$ is trained only on the training set. For each image, Grad-CAM produces a normalised saliency map $S(\mathbf{x})\in[0,1]^{H\times W}$. High- and low-saliency regions are defined as
\begin{equation}
\begin{aligned}
R_h(\mathbf{x}) &= \{(i,j)\mid S(\mathbf{x})_{ij}>\tau_h\},\\
R_b(\mathbf{x}) &= \{(i,j)\mid S(\mathbf{x})_{ij}\leq\tau_b\},
\end{aligned}
\label{eq:regions}
\end{equation}
where $\tau_h$ and $\tau_b$ are high- and low-saliency thresholds. The mixing mask is restricted to overlapping low-saliency regions:
\begin{equation}
M_{ij}=\mathbb{1}\left[(i,j)\in R_b(\mathbf{x}_1)\cap R_b(\mathbf{x}_2)\right].
\label{eq:mask}
\end{equation}
This prevents direct replacement of regions considered diagnostically important in either source image.

\subsection{Adaptive saliency-preserving mixing}
To reduce abrupt transitions, the binary mask is smoothed using a Gaussian kernel $G_\sigma$:
\begin{equation}
M_s=G_\sigma * M.
\label{eq:smoothmask}
\end{equation}
The augmented image is generated by spatially weighted blending:
\begin{equation}
\tilde{\mathbf{x}}=(1-\alpha M_s)\odot \mathbf{x}_1+\alpha M_s\odot \mathbf{x}_2,
\label{eq:mixing}
\end{equation}
where $\odot$ denotes element-wise multiplication and $\alpha$ is an adaptive mixing ratio. Unlike global Mixup, this formulation preserves high-saliency regions while allowing controlled variation in low-saliency areas.

The mixing ratio is determined from saliency similarity and background overlap:
\begin{equation}
\rho = 1 - \frac{|\bar{S}_1-\bar{S}_2|}{\bar{S}_1+\bar{S}_2+\epsilon},\qquad
\eta = \frac{|R_b^{(1)}\cap R_b^{(2)}|}{|R_b^{(1)}\cup R_b^{(2)}|},
\label{eq:rhoeta}
\end{equation}
where $\bar{S}_1$ and $\bar{S}_2$ are the mean saliency values in the background regions of $\mathbf{x}_1$ and $\mathbf{x}_2$. The adaptive ratio is
\begin{equation}
\alpha = \operatorname{clip}\left(\alpha_{\mathrm{base}}\cdot\rho\cdot\eta\cdot\omega,\alpha_{\min},\alpha_{\max}\right),
\label{eq:alpha}
\end{equation}
where $\omega$ is a modality-dependent conservativeness factor. A smaller value is used for radiographs because subtle opacity and contour information may be diagnostically relevant.

\subsection{Diffusion-guided local refinement}
MedDiffuseMix refines only the low-saliency mixed regions rather than synthesising the full image. Gaussian noise is introduced according to
\begin{equation}
q(\mathbf{z}_t\mid \mathbf{z}_{t-1})=
\mathcal{N}\left(\sqrt{1-\beta_t}\mathbf{z}_{t-1},\beta_t\mathbf{I}\right),
\label{eq:forwarddiff}
\end{equation}
where $\beta_t$ is the noise schedule. The reverse denoising process estimates clean content within the low-saliency mask, while high-saliency pixels from $\mathbf{x}_1$ are preserved. This local refinement reduces unrealistic blending artefacts and introduces texture diversity without relying on full-image hallucination.

\subsection{Saliency-preservation constraint}
After augmentation, the guidance classifier produces a saliency map $S(\tilde{\mathbf{x}})$. The augmented image is accepted only if the mean saliency within the original high-saliency region remains above a threshold $\delta$:
\begin{equation}
\frac{1}{|R_h(\mathbf{x}_1)|}
\sum_{(i,j)\in R_h(\mathbf{x}_1)}
S(\tilde{\mathbf{x}})_{ij} \geq \delta.
\label{eq:constraint}
\end{equation}
If this constraint is violated, $\alpha$ is reduced and the augmented image is recomputed. This step ensures that the augmentation preserves the region-level evidence supporting the class label.

\begin{figure*}[t]
\centering
\includegraphics[width=\linewidth]{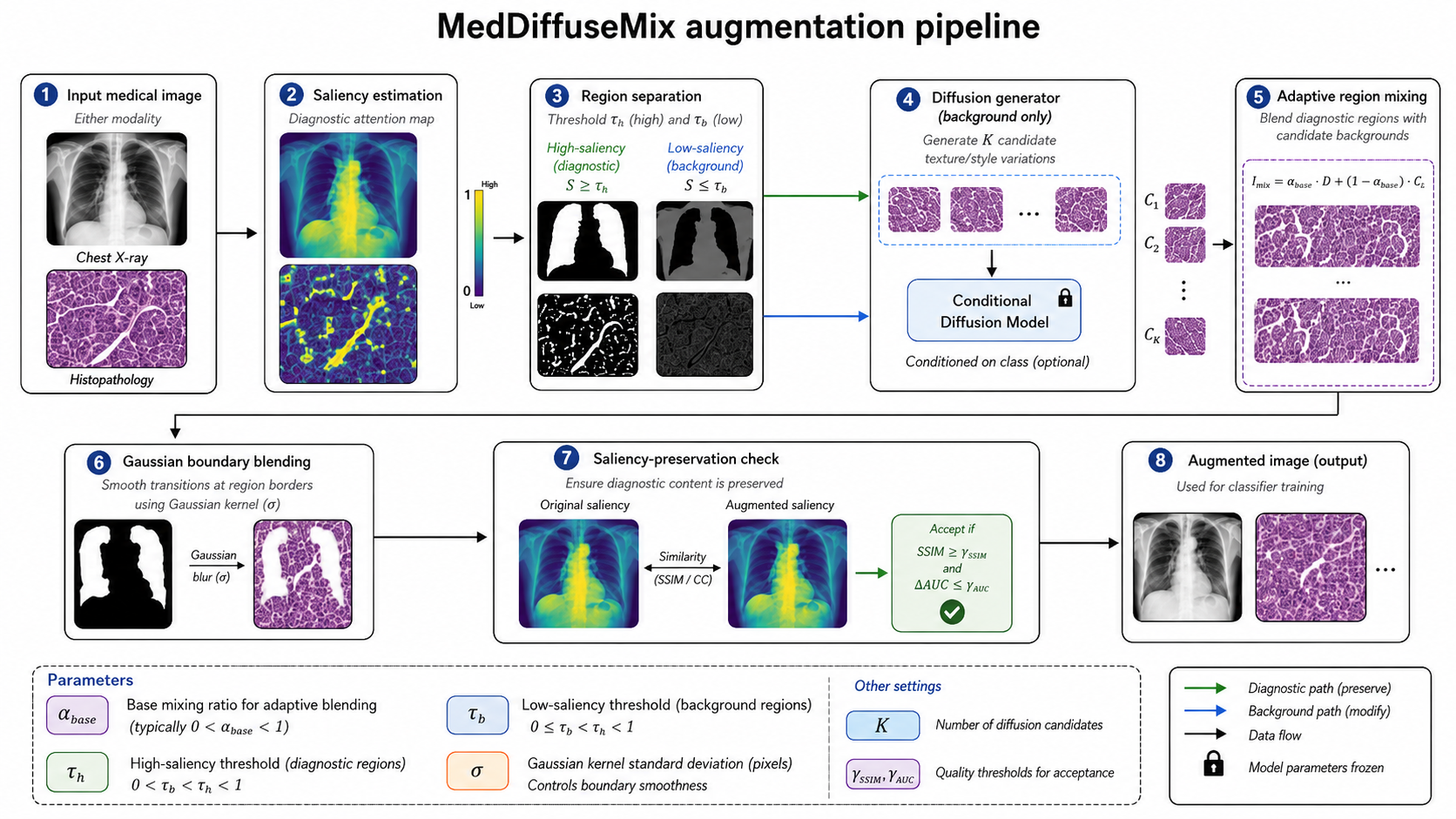}
\caption{Overview of the MedDiffuseMix training-time augmentation pipeline. Saliency maps identify diagnostic regions to preserve, while overlapping low-saliency regions are modified through adaptive mixing, Gaussian boundary smoothing, diffusion-guided local refinement, and saliency-preservation checking. The augmented image is used only during training; inference-time architecture and cost remain unchanged.}
\label{fig:Architecture}
\end{figure*}

\begin{algorithm}[t]
\caption{MedDiffuseMix Augmentation}
\label{alg:meddiffusemix_augmentation}
\begin{algorithmic}[1]
\Require Same-class image pair $\mathbf{x}_1,\mathbf{x}_2$, guidance model $g_\phi$, thresholds $\tau_h,\tau_b,\delta$, base ratio $\alpha_{\mathrm{base}}$, Gaussian parameter $\sigma$
\Ensure Augmented image $\tilde{\mathbf{x}}$
\State $S_1 \gets \mathrm{GradCAM}(g_\phi,\mathbf{x}_1)$; \quad $S_2 \gets \mathrm{GradCAM}(g_\phi,\mathbf{x}_2)$
\State Define $R_h^{(1)}$, $R_b^{(1)}$, and $R_b^{(2)}$ using Eq.~\ref{eq:regions}
\State Construct $M$ using Eq.~\ref{eq:mask}
\State Smooth mask: $M_s \gets G_\sigma * M$
\State Compute $\rho$, $\eta$, and $\alpha$ using Eqs.~\ref{eq:rhoeta}--\ref{eq:alpha}
\State Generate $\tilde{\mathbf{x}}$ using Eq.~\ref{eq:mixing}
\State Apply diffusion-guided refinement within the low-saliency mask
\State Compute $S(\tilde{\mathbf{x}})\gets \mathrm{GradCAM}(g_\phi,\tilde{\mathbf{x}})$
\While{$\frac{1}{|R_h^{(1)}|}\sum_{(i,j)\in R_h^{(1)}}S(\tilde{\mathbf{x}})_{ij}<\delta$ and $\alpha>\alpha_{\min}$}
    \State $\alpha \gets 0.8\alpha$
    \State Recompute $\tilde{\mathbf{x}}$ using Eq.~\ref{eq:mixing}
    \State Update $S(\tilde{\mathbf{x}})\gets \mathrm{GradCAM}(g_\phi,\tilde{\mathbf{x}})$
\EndWhile
\State \Return $\tilde{\mathbf{x}}$
\end{algorithmic}
\end{algorithm}

\section{Experimental setup}
\label{sec:experiments}
The evaluation uses four public medical imaging benchmarks covering radiography and histopathology: RSNA Pneumonia, MURA Fracture, PatchCamelyon, and BreakHis. Experiments are reported across five classifier backbones: ResNet-50, DenseNet-121, EfficientNet-B4, ViT-B/16, and Swin-T. MedDiffuseMix is compared with no augmentation, standard augmentation, Mixup, GenMix, SaliencyMix, and diffusion-based augmentation. Performance is evaluated using accuracy, precision, recall, F1-score, and AUC, with mean and standard deviation reported where applicable. Qualitative augmentation examples, ablation analysis, hyperparameter sensitivity, and Grad-CAM visualisation are used to assess whether the method preserves diagnostically relevant evidence while improving training diversity.

\section{Results}
\label{sec:results}

\subsection{Overall classification performance}
Table~\ref{tab:overall_results} summarises accuracy across four datasets, five classifier backbones, and seven augmentation settings. MedDiffuseMix obtains the highest reported accuracy in every dataset-backbone combination. The improvement is not confined to a single architecture: gains are observed for convolutional networks and transformer-based models, suggesting that the method acts as a training-data regulariser rather than as a backbone-specific tuning effect.

\begin{table*}[t]
\centering
\scriptsize
\setlength{\tabcolsep}{4pt}
\caption{Overall accuracy (\%, mean \(\pm\) std) comparison of augmentation methods across datasets and model architectures.}
\label{tab:overall_results}
\begin{adjustbox}{max width=\textwidth}
\begin{tabular}{@{} ll *{7}{c} @{}}
\toprule
\textbf{Dataset} & \textbf{Model} & \textbf{No Aug} & \textbf{Standard Aug} & \textbf{Mixup \cite{ref42}} & \textbf{GenMix \cite{ref21}} & \textbf{SaliencyMix \cite{ref43}} & \textbf{DBA \cite{ref1}} & \textbf{MedDiffuseMix} \\
\midrule

\multirow{5}{*}{RSNA Pneumonia}
& ResNet-50 & \(85.2 \pm 0.4\) & \(86.1 \pm 0.3\) & \(86.8 \pm 0.4\) & \(87.2 \pm 0.3\) & \(87.5 \pm 0.3\) & \(87.9 \pm 0.4\) & \textbf{\(89.3 \pm 0.2\)} \\
& DenseNet-121 & \(86.4 \pm 0.3\) & \(87.2 \pm 0.2\) & \(87.6 \pm 0.3\) & \(88.1 \pm 0.2\) & \(88.4 \pm 0.3\) & \(88.7 \pm 0.3\) & \textbf{\(90.1 \pm 0.2\)} \\
& EfficientNet-B4 & \(87.1 \pm 0.3\) & \(88.0 \pm 0.3\) & \(88.5 \pm 0.3\) & \(89.0 \pm 0.2\) & \(89.3 \pm 0.2\) & \(89.6 \pm 0.3\) & \textbf{\(91.2 \pm 0.1\)} \\
& ViT-B/16 & \(84.8 \pm 0.5\) & \(85.7 \pm 0.4\) & \(86.3 \pm 0.4\) & \(86.8 \pm 0.4\) & \(87.1 \pm 0.3\) & \(87.4 \pm 0.4\) & \textbf{\(88.9 \pm 0.3\)} \\
& Swin-T & \(86.7 \pm 0.3\) & \(87.5 \pm 0.3\) & \(88.0 \pm 0.3\) & \(88.4 \pm 0.3\) & \(88.7 \pm 0.3\) & \(89.0 \pm 0.3\) & \textbf{\(90.5 \pm 0.2\)} \\
\midrule

\multirow{5}{*}{MURA Fracture}
& ResNet-50 & \(84.6 \pm 0.5\) & \(85.3 \pm 0.4\) & \(85.9 \pm 0.4\) & \(86.4 \pm 0.4\) & \(86.7 \pm 0.4\) & \(87.0 \pm 0.4\) & \textbf{\(88.4 \pm 0.3\)} \\
& DenseNet-121 & \(85.8 \pm 0.4\) & \(86.5 \pm 0.3\) & \(87.0 \pm 0.3\) & \(87.4 \pm 0.3\) & \(87.7 \pm 0.3\) & \(88.0 \pm 0.3\) & \textbf{\(89.5 \pm 0.2\)} \\
& EfficientNet-B4 & \(86.5 \pm 0.4\) & \(87.3 \pm 0.3\) & \(87.8 \pm 0.3\) & \(88.3 \pm 0.3\) & \(88.6 \pm 0.3\) & \(88.9 \pm 0.3\) & \textbf{\(90.6 \pm 0.2\)} \\
& ViT-B/16 & \(84.2 \pm 0.6\) & \(85.0 \pm 0.5\) & \(85.6 \pm 0.5\) & \(86.0 \pm 0.5\) & \(86.3 \pm 0.5\) & \(86.6 \pm 0.5\) & \textbf{\(88.0 \pm 0.4\)} \\
& Swin-T & \(85.9 \pm 0.4\) & \(86.7 \pm 0.4\) & \(87.2 \pm 0.4\) & \(87.6 \pm 0.4\) & \(87.9 \pm 0.4\) & \(88.2 \pm 0.4\) & \textbf{\(89.7 \pm 0.3\)} \\
\midrule

\multirow{5}{*}{PatchCamelyon}
& ResNet-50 & \(84.8 \pm 0.3\) & \(85.6 \pm 0.3\) & \(86.1 \pm 0.3\) & \(86.6 \pm 0.3\) & \(86.9 \pm 0.3\) & \(87.2 \pm 0.3\) & \textbf{\(88.7 \pm 0.2\)} \\
& DenseNet-121 & \(85.9 \pm 0.2\) & \(86.7 \pm 0.2\) & \(87.2 \pm 0.2\) & \(87.6 \pm 0.2\) & \(87.9 \pm 0.2\) & \(88.2 \pm 0.2\) & \textbf{\(89.8 \pm 0.1\)} \\
& EfficientNet-B4 & \(86.7 \pm 0.2\) & \(87.5 \pm 0.2\) & \(88.0 \pm 0.2\) & \(88.4 \pm 0.2\) & \(88.7 \pm 0.2\) & \(89.0 \pm 0.2\) & \textbf{\(90.9 \pm 0.1\)} \\
& ViT-B/16 & \(84.4 \pm 0.4\) & \(85.2 \pm 0.4\) & \(85.8 \pm 0.4\) & \(86.2 \pm 0.4\) & \(86.5 \pm 0.4\) & \(86.8 \pm 0.4\) & \textbf{\(88.3 \pm 0.3\)} \\
& Swin-T & \(86.2 \pm 0.3\) & \(87.0 \pm 0.3\) & \(87.5 \pm 0.3\) & \(87.9 \pm 0.3\) & \(88.2 \pm 0.3\) & \(88.5 \pm 0.3\) & \textbf{\(90.0 \pm 0.2\)} \\
\midrule

\multirow{5}{*}{BreakHis}
& ResNet-50 & \(87.3 \pm 0.4\) & \(88.1 \pm 0.3\) & \(88.6 \pm 0.3\) & \(89.1 \pm 0.3\) & \(89.4 \pm 0.3\) & \(89.7 \pm 0.3\) & \textbf{\(91.5 \pm 0.2\)} \\
& DenseNet-121 & \(88.4 \pm 0.3\) & \(89.2 \pm 0.2\) & \(89.7 \pm 0.2\) & \(90.1 \pm 0.2\) & \(90.4 \pm 0.2\) & \(90.7 \pm 0.2\) & \textbf{\(92.6 \pm 0.1\)} \\
& EfficientNet-B4 & \(89.2 \pm 0.3\) & \(90.0 \pm 0.2\) & \(90.5 \pm 0.2\) & \(90.9 \pm 0.2\) & \(91.2 \pm 0.2\) & \(91.5 \pm 0.2\) & \textbf{\(93.4 \pm 0.1\)} \\
& ViT-B/16 & \(87.0 \pm 0.5\) & \(87.8 \pm 0.4\) & \(88.3 \pm 0.4\) & \(88.8 \pm 0.4\) & \(89.1 \pm 0.4\) & \(89.4 \pm 0.4\) & \textbf{\(91.2 \pm 0.3\)} \\
& Swin-T & \(88.7 \pm 0.3\) & \(89.5 \pm 0.3\) & \(90.0 \pm 0.3\) & \(90.4 \pm 0.3\) & \(90.7 \pm 0.3\) & \(91.0 \pm 0.3\) & \textbf{\(92.9 \pm 0.2\)} \\
\bottomrule
\end{tabular}
\end{adjustbox}
\end{table*}

\begin{figure}
    \centering
    \includegraphics[width=\linewidth]{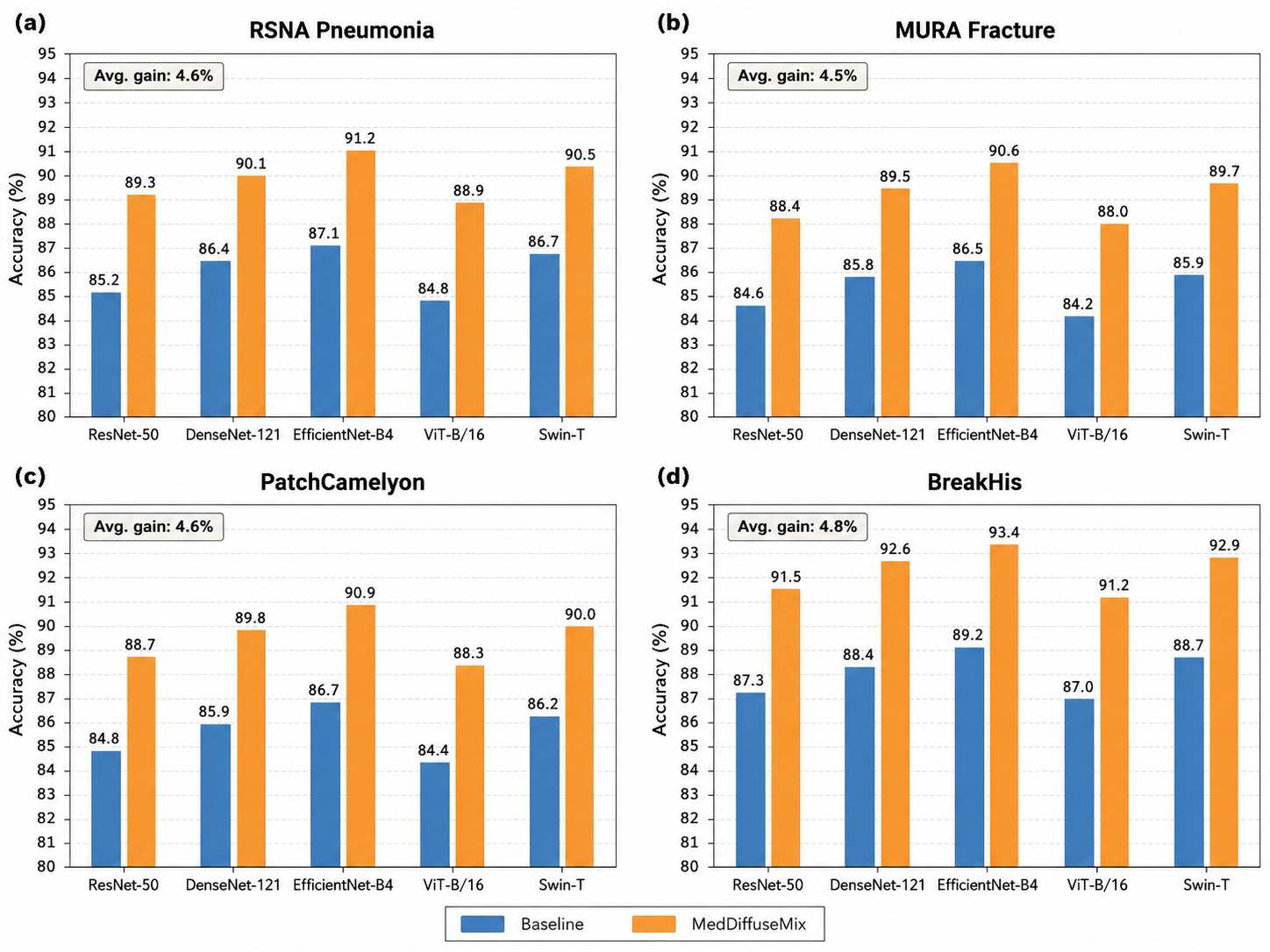}
    \caption{Overall classification accuracy across four datasets and five backbone architectures. Each panel compares the unaugmented baseline with MedDiffuseMix; the annotated callout reports the average gain within each dataset.}
    \label{fig:graphresult}
\end{figure}

\subsection{Discrimination and class-wise balance}
For the best-performing backbone in the reported experiments, EfficientNet-B4, Table~\ref{tab:detailed_metrics} reports precision, recall, F1-score, and AUC. MedDiffuseMix improves all four metrics over the strongest listed baseline on each dataset. This is important for medical imaging because an augmentation method that improves accuracy while reducing recall or AUC would be less useful for screening and triage settings.

\begin{table*}[htbp]
\centering
\caption{Detailed performance metrics for EfficientNet-B4 across datasets with different augmentation methods.}
\label{tab:detailed_metrics}
\begin{tabular}{@{}llcccc@{}}
\toprule
\textbf{Dataset} & \textbf{Augmentation} & \textbf{Precision} & \textbf{Recall} & \textbf{F1-Score} & \textbf{AUC} \\
\midrule
\multirow{7}{*}{RSNA Pneumonia}
& No Augmentation & 87.3 & 86.9 & 87.1 & 0.934 \\
& Standard Aug & 88.2 & 87.8 & 88.0 & 0.942 \\
& Mixup & 88.7 & 88.3 & 88.5 & 0.947 \\
& GenMix & 89.2 & 88.8 & 89.0 & 0.951 \\
& SaliencyMix & 89.5 & 89.1 & 89.3 & 0.954 \\
& DBA & 89.8 & 89.4 & 89.6 & 0.956 \\
& \textbf{MedDiffuseMix} & \textbf{91.4} & \textbf{91.0} & \textbf{91.2} & \textbf{0.968} \\
\midrule

\multirow{7}{*}{MURA Fracture}
& No Augmentation & 86.7 & 86.3 & 86.5 & 0.928 \\
& Standard Aug & 87.5 & 87.1 & 87.3 & 0.936 \\
& Mixup & 88.0 & 87.6 & 87.8 & 0.940 \\
& GenMix & 88.5 & 88.1 & 88.3 & 0.944 \\
& SaliencyMix & 88.8 & 88.4 & 88.6 & 0.947 \\
& DBA & 89.1 & 88.7 & 88.9 & 0.949 \\
& \textbf{MedDiffuseMix} & \textbf{90.8} & \textbf{90.4} & \textbf{90.6} & \textbf{0.962} \\
\midrule

\multirow{7}{*}{PatchCamelyon}
& No Augmentation & 86.9 & 86.5 & 86.7 & 0.930 \\
& Standard Aug & 87.7 & 87.3 & 87.5 & 0.938 \\
& Mixup & 88.2 & 87.8 & 88.0 & 0.942 \\
& GenMix & 88.6 & 88.2 & 88.4 & 0.946 \\
& SaliencyMix & 88.9 & 88.5 & 88.7 & 0.948 \\
& DBA & 89.2 & 88.8 & 89.0 & 0.950 \\
& \textbf{MedDiffuseMix} & \textbf{91.1} & \textbf{90.7} & \textbf{90.9} & \textbf{0.965} \\
\midrule

\multirow{7}{*}{BreakHis}
& No Augmentation & 89.4 & 89.0 & 89.2 & 0.945 \\
& Standard Aug & 90.2 & 89.8 & 90.0 & 0.953 \\
& Mixup & 90.7 & 90.3 & 90.5 & 0.957 \\
& GenMix & 91.1 & 90.7 & 90.9 & 0.960 \\
& SaliencyMix & 91.4 & 91.0 & 91.2 & 0.963 \\
& DBA & 91.7 & 91.3 & 91.5 & 0.965 \\
& \textbf{MedDiffuseMix} & \textbf{93.6} & \textbf{93.2} & \textbf{93.4} & \textbf{0.978} \\
\bottomrule
\end{tabular}
\end{table*}

\subsection{Qualitative augmentation behaviour}
\begin{figure}
    \centering
    \includegraphics[width=\linewidth]{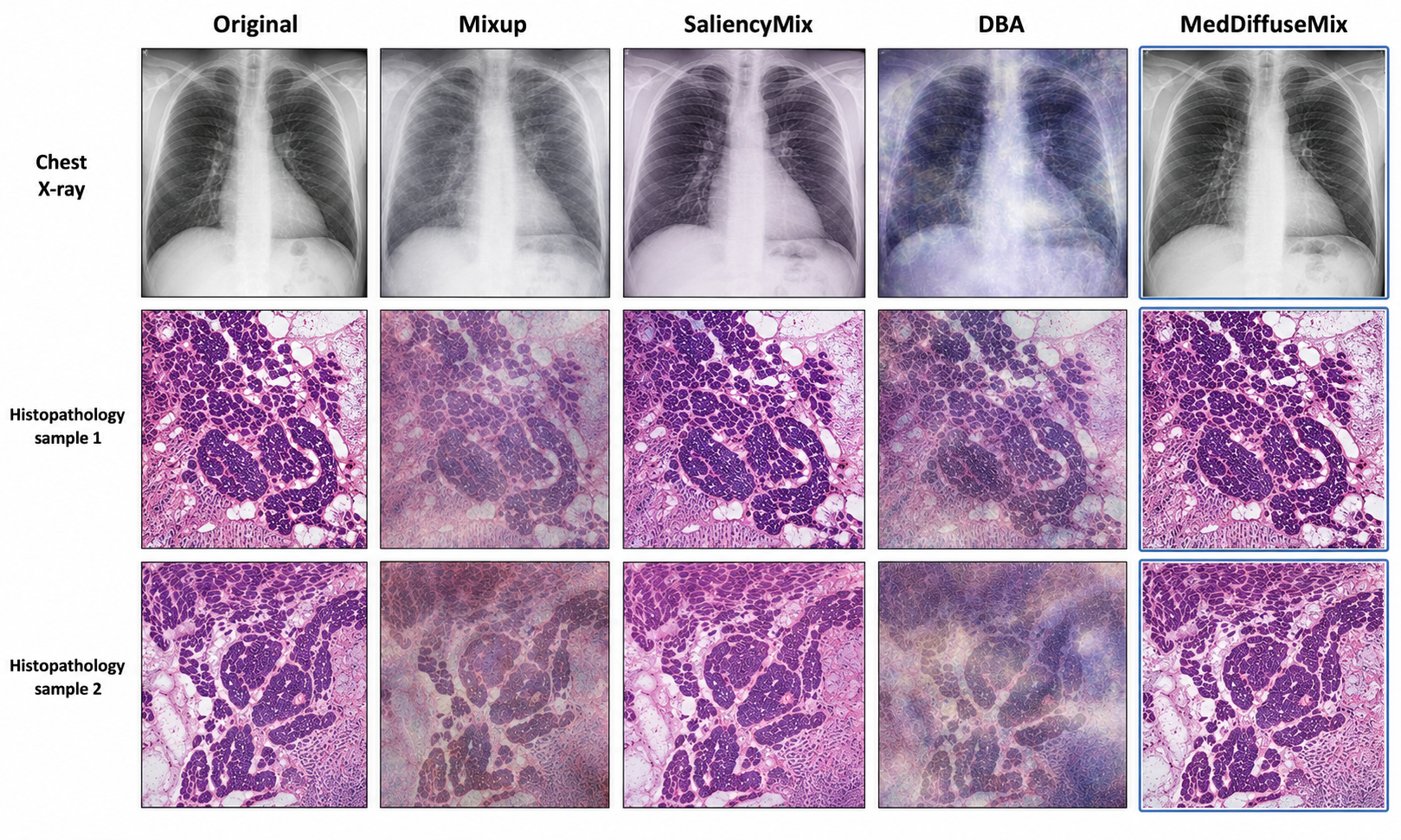}
    \caption{Qualitative comparison of augmentation methods on chest X-ray (top row) and histopathology (bottom row) images. From left to right: Original, Mixup, SaliencyMix, DBA, and the proposed MedDiffuseMix. Mixup introduces global intensity blending that can dilute diagnostic structures, while SaliencyMix preserves salient regions but may introduce abrupt transitions. DBA/DiffuseMix improves texture diversity but may alter clinically relevant areas. MedDiffuseMix is designed to concentrate augmentation in low-saliency regions while preserving diagnostically salient regions identified by the guidance model.}
    \label{fig:augmented}
\end{figure}

Figure~\ref{fig:augmented} compares representative augmented samples. Standard mixing methods can introduce global blending or abrupt boundaries, and unconstrained generative augmentation can alter regions that may contain diagnostic information. MedDiffuseMix is designed to limit such changes by masking high-saliency regions and smoothing low-saliency transitions.

\subsection{Ablation and sensitivity analysis}
\begin{figure}[htbp]
    \centering
    \includegraphics[width=\linewidth]{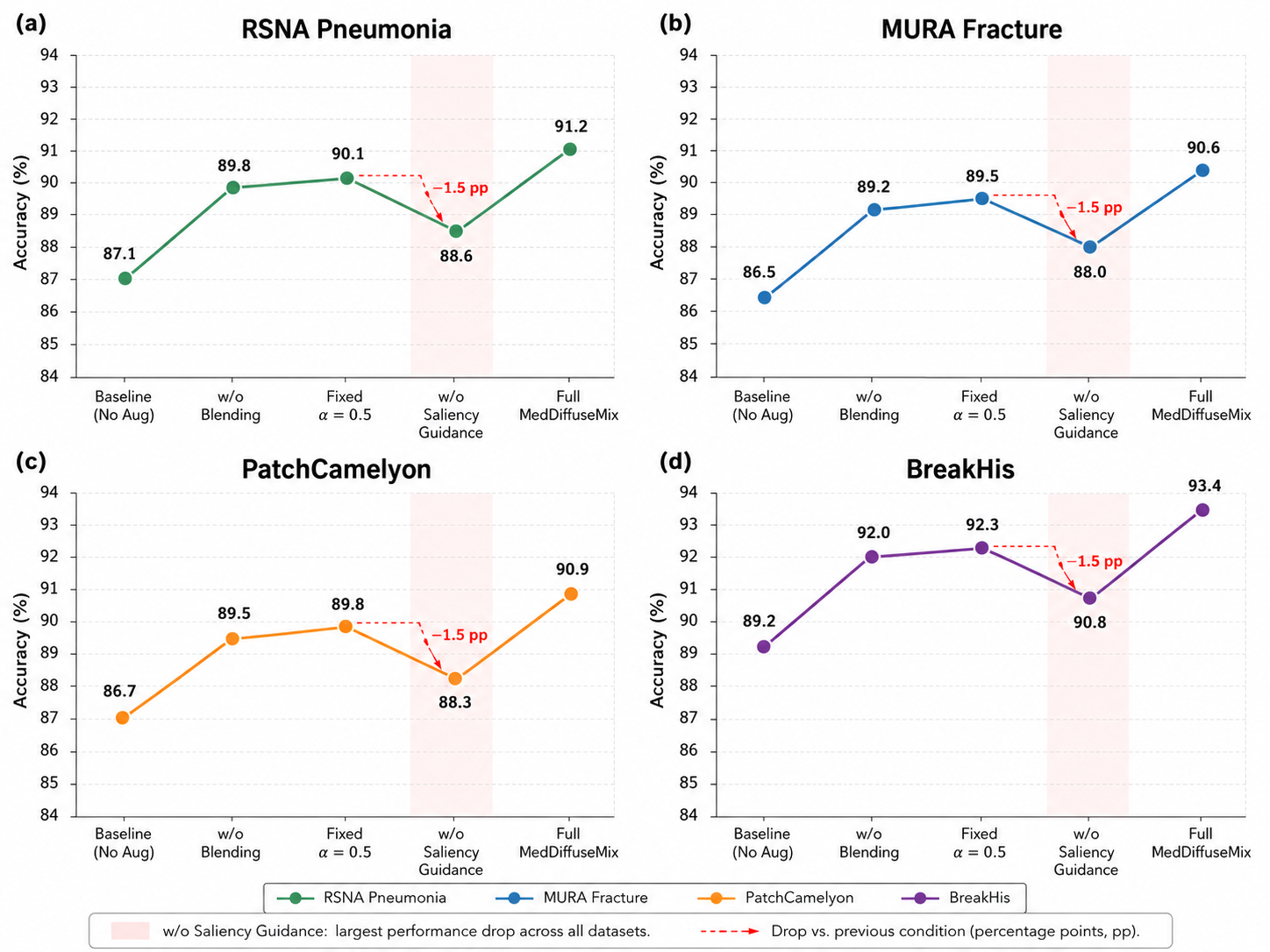}
    \caption{Ablation analysis across four datasets using EfficientNet-B4. The full MedDiffuseMix configuration performs best on every dataset, while removing saliency guidance yields the largest drop, indicating that diagnostic-region preservation is the most critical component.}
    \label{fig:ablation}
\end{figure}

\begin{figure}[htbp]
    \centering
    \includegraphics[width=\linewidth]{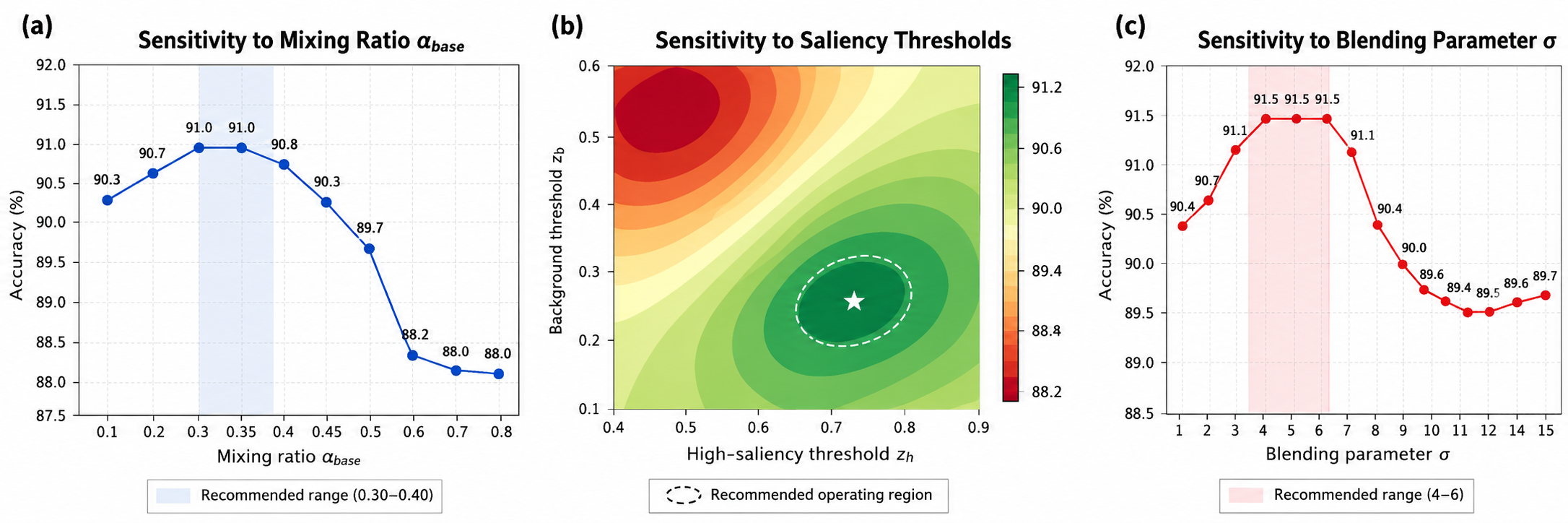}
\caption{Sensitivity analysis of the main MedDiffuseMix hyperparameters on the RSNA Pneumonia dataset. Panel (a) analyses the base mixing ratio $\alpha_{\mathrm{base}}$, panel (b) visualises the interaction between the high-saliency and background thresholds, and panel (c) analyses the Gaussian blending parameter $\sigma$. Shaded regions indicate empirically stable operating ranges.}
    \label{fig:parameter_sensitivity}
\end{figure}

The ablation study in Figure~\ref{fig:ablation} evaluates the contribution of individual components. Removing saliency guidance directly weakens diagnostic preservation; removing adaptive mixing reduces flexibility across modalities; removing boundary smoothing increases the risk of artefactual transitions. Hyperparameter sensitivity in Figure~\ref{fig:parameter_sensitivity} suggests that conservative mixing is preferable for radiographs, whereas histopathology tolerates stronger texture variation.

\section{Explainability and diagnostic preservation}
\label{sec:explainability}

Explainability analysis is used here for two purposes: to guide augmentation and to audit whether the augmented samples preserve the model's diagnostic attention. Grad-CAM heatmaps are computed for original and augmented images. The goal is not to claim that Grad-CAM fully captures clinical reasoning, but to check whether augmentation moves the classifier's attention away from plausible disease-relevant regions.

\begin{figure}[htbp]
\centering
\includegraphics[width=\linewidth]{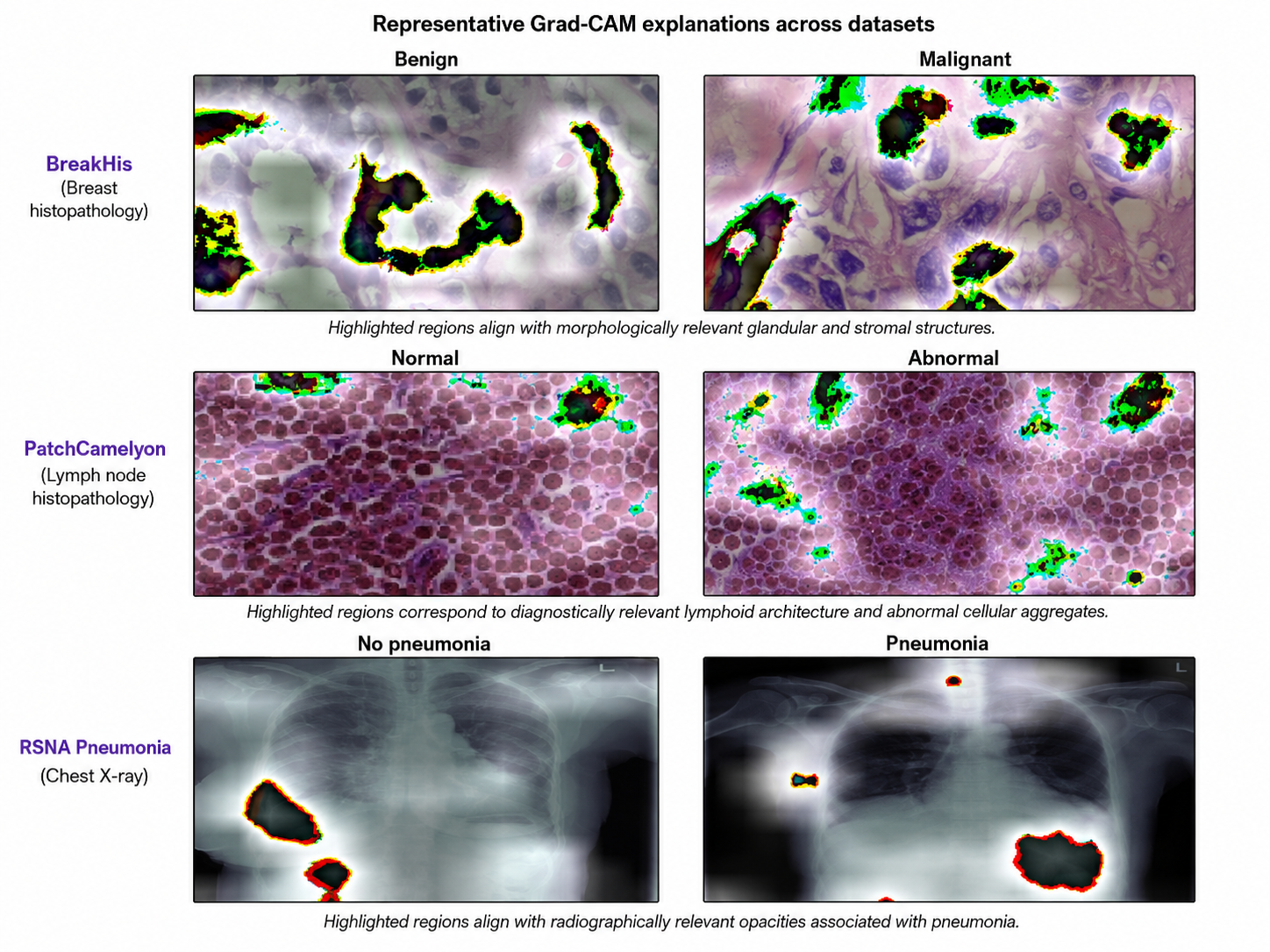}
\caption{Representative Grad-CAM explanations across the BreakHis, PatchCamelyon, and RSNA Pneumonia datasets. Rows correspond to datasets and columns correspond to paired negative/benign and positive/abnormal examples. The highlighted regions remain aligned with diagnostically relevant structures, supporting the use of saliency as both a guidance signal and a post-augmentation audit mechanism.}
\label{fig:gradcam_examples}
\end{figure}

Across the visual examples, the main qualitative pattern is that high-saliency image evidence remains spatially similar after MedDiffuseMix augmentation. This behaviour is expected from Eq.~\ref{eq:constraint}, which reduces the mixing strength when the augmented image shifts the attention map away from the original high-saliency region.

\begin{figure}
    \centering
    \includegraphics[width=\linewidth]{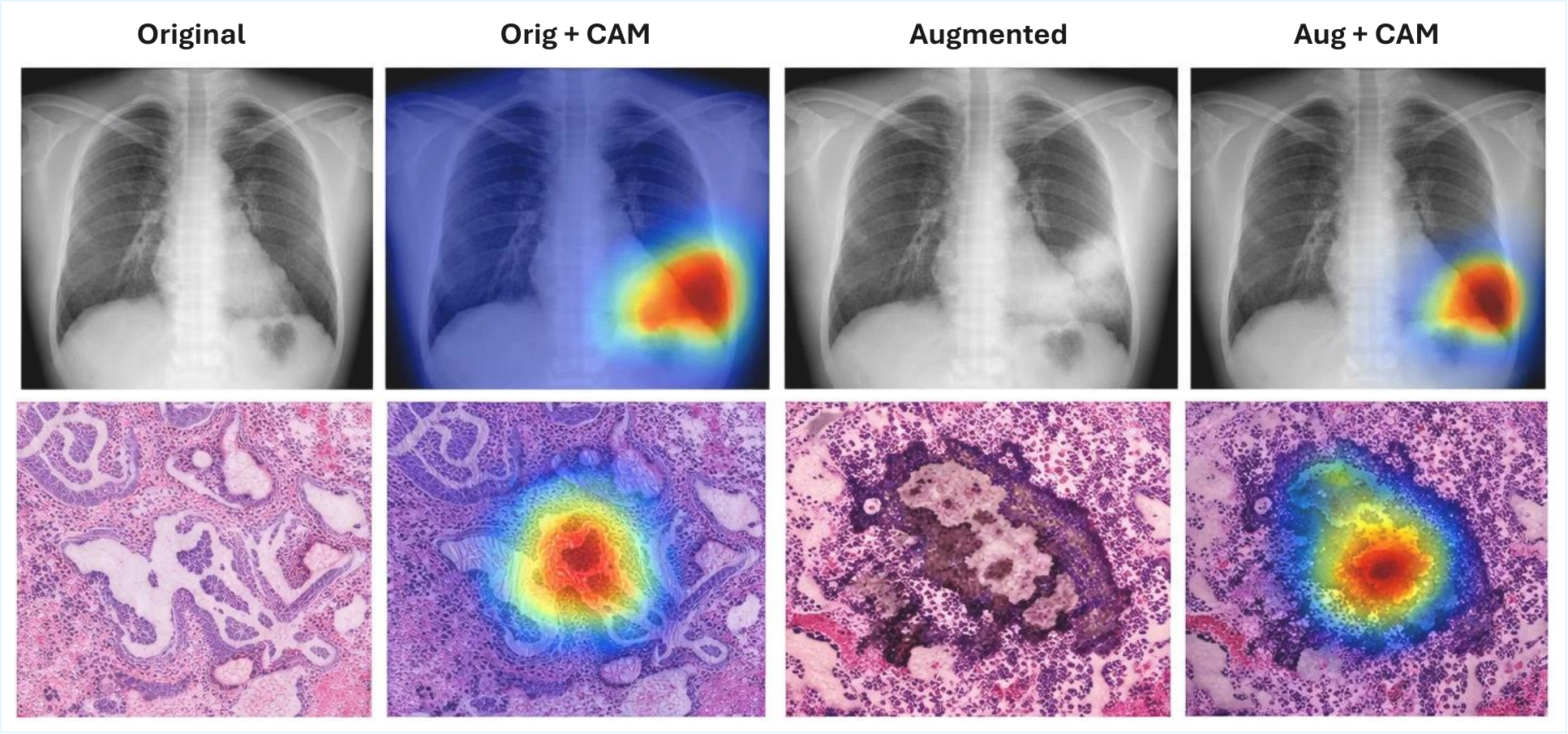}
    \caption{Saliency preservation during augmentation. For each sample, the first row shows a chest X-ray and the second row shows a histopathology patch. Columns correspond to the original image, original with Grad-CAM, MedDiffuseMix augmented image, and augmented image with Grad-CAM. High-saliency diagnostic regions remain spatially aligned before and after augmentation, providing visual support for the saliency-preservation constraint.}
    \label{fig:Saliency}
\end{figure}

This saliency analysis should be interpreted as a model-auditing tool rather than clinical proof. Stronger validation would require expert reader assessment of augmented-image plausibility, lesion-presence agreement, and blinded review of whether augmented samples preserve clinically meaningful findings.

\section{Conclusion}
\label{sec:conclusion}

This paper presented MedDiffuseMix, a saliency-preserving diffusion augmentation framework for limited-data medical image classification. By combining classifier-derived saliency maps, low-saliency region mixing, diffusion-guided local refinement, adaptive mixing, and saliency-preservation checking, the method increases appearance diversity while reducing the risk of corrupting diagnostic evidence. Across four public benchmarks and five classifier backbones, MedDiffuseMix achieved consistent improvements over standard and advanced augmentation baselines. Ablation, sensitivity, and attribution analyses indicate that these gains are associated with preserving diagnostically salient regions rather than introducing uncontrolled synthetic artefacts.

There are several limitation, the evaluation is restricted to 2D classification on public datasets and stratified subsets; broader external validation is required across institutions, scanners, acquisition protocols, and patient subgroups. The method also depends on Grad-CAM and an initial guidance classifier, which may produce unstable or biased saliency estimates. Future work should extend MedDiffuseMix to volumetric CT and MRI, 3D pathology workflows, lesion-aware segmentation, and weakly supervised whole-slide analysis, while incorporating alternative attribution signals, expert review, public code, fixed splits, and external validation cohorts.

\section*{Ethics statement}
This study used publicly available datasets and did not involve new patient recruitment or direct interaction with human participants.

\section*{Data and code availability}
The datasets analysed in this study are publicly available from their respective repositories. The implementation of MedDiffuseMix is available at: \url{https://github.com/rajavavek/MedDiffuseMix}.

\section*{Author contributions}
T.K. proposed the MedDiffuseMix framework, designed the diffusion-based augmentation algorithm, performed theoretical analysis, and wrote the initial manuscript. R.V. curated and preprocessed datasets, conducted experiments, validated results, and revised the manuscript. M.T. contributed to methodology refinement, interpretation of results, and manuscript review and editing. All authors verified and approved the final submission.

\section*{Funding}
This work has received no funding.

\section*{Declaration of competing interest}
The authors declare no competing interests.

\bibliographystyle{elsarticle-num}
\bibliography{sample}

\end{document}